\documentclass[10pt,twocolumn,letterpaper]{article}

\usepackage[pagenumbers]{cvpr} 

\usepackage[dvipsnames]{xcolor}

\usepackage{tikz}

\usepackage[utf8]{inputenc} 
\usepackage[T1]{fontenc}    
\usepackage{url}            
\usepackage{booktabs}       
\usepackage{amsfonts}       
\usepackage{nicefrac}       
\usepackage{microtype}      
\usepackage{xcolor}         
\usepackage{graphicx}
\usepackage[export]{adjustbox}
\usepackage{mathtools}

\usepackage[normalem]{ulem}
\useunder{\uline}{\ul}{}

\usepackage{array}
\newcolumntype{H}{>{\setbox0=\hbox\bgroup}c<{\egroup}@{}}
\usepackage{float}
\usepackage[skip=5pt, belowskip=-10pt]{caption}

\definecolor{cvprblue}{rgb}{0.21,0.49,0.74}
\usepackage[pagebackref,breaklinks,colorlinks,citecolor=cvprblue]{hyperref}

\def\paperID{*****} 
\def\confName{CVPR}
\def\confYear{2024}

\title{DocSynthv2: A Practical Autoregressive Modeling for Document Generation}

\newcommand*{\affaddr}[1]{#1} 
\newcommand*{\affmark}[1][*]{\textsuperscript{#1}}
\newcommand*{\email}[1]{\texttt{#1}}
\author{
 Sanket Biswas\affmark[1]\thanks{Work done during internship at Adobe Research}\and
 Rajiv Jain\affmark[2]\and
 Vlad I. Morariu\affmark[2]\and
 Jiuxiang Gu\affmark[2]\and 
 Puneet Mathur\affmark[2]\and
 Curtis Wigington\affmark[2]\and
 Tong Sun\affmark[2]\and
 Josep Llad\'{o}s\affmark[1]\and \\
\affaddr{\affmark[1] Computer Vision Center, UAB,
Spain}
\affaddr{\affmark[2] Adobe Research}\\
\email{\tt\small \{sbiswas, josep\}@cvc.uab.es}
\email{\tt\small \{rajijain, morariu, jigu, puneetm, wigingto, tsun\}@adobe.com}
}

\begin{document}
\maketitle
\begin{abstract}

While the generation of document layouts has been extensively explored, comprehensive document generation—encompassing both layout and content—presents a more complex challenge. This paper delves into this advanced domain, proposing a novel approach called \textbf{DocSynthv2} through the development of a simple yet effective autoregressive structured model. Our model, distinct in its integration of both layout and textual cues, marks a  step beyond existing layout-generation approaches. By focusing on the  relationship between the structural elements and the textual content within documents, we aim to generate cohesive and contextually relevant documents without any reliance on visual components. Through experimental studies on our curated benchmark for the new task, we demonstrate the ability of our model combining layout and textual information in enhancing the generation quality and relevance of documents, opening new pathways for research in document creation and automated design. Our findings emphasize the effectiveness of autoregressive models in handling complex document generation tasks.

\end{abstract}    
\section{Introduction}
\label{sec:intro}






Recent advancements in generative models~\cite{zhang2023layoutdiffusion, zheng2023layoutdiffusion, esser2021taming, brown2020language} have made significant impacts on language, image, and multimodal content generation. There is an increasing focus on vector graphic document generation~\cite{inoue2023towards, shimoda2024towards, yamaguchi2021canvasvae} within this realm, where these models support users in creating, modifying, publishing, and designing both business and artistic documents. Documents differ from standard natural images as they contain structured layers of text and media content. The field of document generation presents unique challenges in seamlessly integrating visual elements such as style, layout, and multimedia with textual content, posing new problems for the vision community. 

Document layout generation~\cite{inoue2023layoutdm,kong2022blt,arroyo2021variational,hui2023unifying,patil2020read,jyothi2019layoutvae,gupta2021layouttransformer} has played a crucial role in numerous applications, ranging from automated report creation to dynamic webpage design, significantly impacting how information is perceived and interacted with by users. With large language models (LLMs)~\cite{brown2020language, touvron2023llama} becoming more and more capable of compositional reasoning of visual concepts~\cite{feng2024layoutgpt}, it opens further avenues for exploiting autoregressive approaches in the automatic end-to-end generation of both document content and layout structure. Moreover, synthetic document generation~\cite{biswas2021docsynth, pisaneschi2023automatic} has gained attention in recent times owing to lack of multi-domain large-scaled layout annotated datasets necessary for document pre-training~\cite{kim2022ocr}. However, end-to-end pixel-based approaches~\cite{biswas2021docsynth, yim2021synthtiger} suffer from low-resolution generated outputs where the textual content can be hardly extracted. In this work, we introduce \textbf{DocSynthv2} to seamlessly generate layout structure with integrated text, essential to convey specific information and context, completing the communication objective of the document.

This work contributes  to document generation research in three different folds: 1) We curate a large-scale extended benchmark called \textbf{PubGenNet} tailor-made for document generation and completion task. 2) We introduce a simple and flexible  autoregressive approach for generating high-resolution document outputs, capable of handling sequences of arbitrary lengths. 3) We outline future challenges and opportunities in evaluating document generation, setting the stage for advancements in this evolving field. 

\section{Related Work}
\label{sec:related}

\begin{figure*}[t]
  \centering
  \includegraphics[width=0.8\linewidth, keepaspectratio]{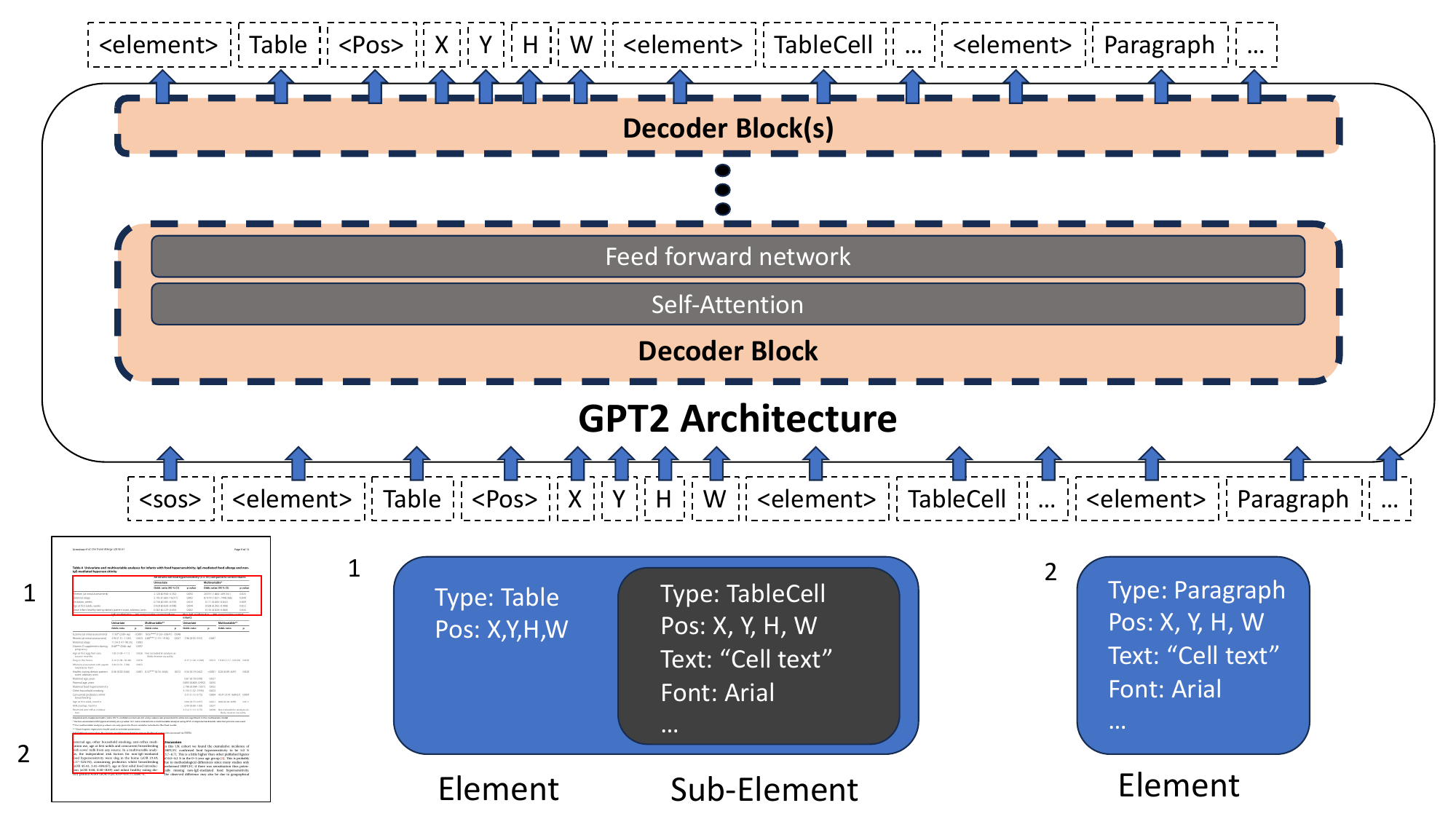}
  \caption{\textbf{Overall Architecture of \texttt{DocSynthv2}}. }
  \label{fig:architecture}
\end{figure*}

\noindent
\textbf{Document Layout Generation}
Recently, there has been a surge in research on layout generation. Foundational works like LayoutGAN~\cite{li2019layoutgan} and LayoutVAE~\cite{jyothi2019layoutvae} have been influential in synthesizing layouts by modeling geometric relations of different 2D elements and then rendering them in the image space. Document layout generation has received extensive interest in recent years owing to its integration in tasks such as content generation~\cite{yamaguchi2021canvasvae, zheng2019content} and graphic web designs~\cite{deka2017rico}. While ~\cite{zheng2019content} attempts to generate document layouts given user-conditioned prompts (eg. input reference image, keywords, and category of the document), ~\cite{patil2020read} proposed an approach to construct hierarchies of document layouts and later sample and generate them using a recursive VAE. The aforementioned method was further extended using
graph autoencoder networks~\cite{patil2021layoutgmn} with optional design constraints for further improvement. Gupta \textit{et. al.}~\cite{gupta2021layouttransformer} proposed \texttt{Layout Transformer} with self-attention~\cite{vaswani2017attention} which is the most relevant to this work. They used a next element prediction objective (i.e. layout completion) using a transformer architecture in an autoregressive manner to produce layout tokens, including class labels and bounding boxes of document objects. ~\cite{arroyo2021variational} tried to combine these generative transformers~\cite{gupta2021layouttransformer} with VAE's to learn better layout refinement and prediction. 

\noindent
\textbf{Synthetic Document Generation} 
The Computer Vision community has also captured emerging interest to generate synthetic realistic scene images with plausible layouts from a user provided reference layout~\cite{zhao2020layout2image, he2021context, johnson2018image}, emphasizing particularly on high-resolution image outputs. The DocSynth framework~\cite{biswas2021docsynth} introduced the first image-to-image translation pipeline for creating synthetic document image datasets for augmenting real data during training for document layout analysis tasks~\cite{pfitzmann2022doclaynet, zhong2019publaynet}. In this work, we move a step forward towards generating synthetic data with content preservation. 

\section{Method}\label{sec:methodology}

In this section, we introduce our proposed approach for the document generation task. We first discuss our representation of document elements essential for model understanding. Next, we discuss the DocSynthv2 framework and show how we can leverage the knowledge of both layout elements and their corresponding content to model the probability distribution of an overall page structure. Lastly, we discuss the learning objectives we have used to train the whole network.  


\subsection{Overview}

\noindent
\textbf{Document Representation} The document layout of a page can comprise multiple sets of elements, where each element can be described by its category $c$, left and top coordinate $x$ and $y$, as well as width $w$ and height $h$. The continuous attributes $x$, $y$, $w$ and $h$ are often quantized, which has proven to be useful for graphic layout generation approaches~\cite{gupta2021layouttransformer, arroyo2021variational, zheng2019content}. Following
the FlexDM approach~\cite{inoue2023towards}, we represent document $\mathcal{D}$ as a vector consisting of a tuple of layout components  $\left(D_1, D_2, \ldots, D_S\right)$, where $S$ is the number of elements in $\mathcal{D}$. Each element $D_i=\left\{d_i^k \mid k \in \mathcal{E}\right\}$ can represent either element type 
, position, style attributes, or raw text content where $k$ represents the indices of the attributes. Contrary to FlexDM~\cite{inoue2023towards}, we do not use any embeddings in the input sequence but rather use only the element's layout information or its content attributes. We concatenate the layout information along with the text attribute tokens for every element as shown in Equation~\ref{eq:1}. Here, $N$ represents the total number of elements, while $\langle\operatorname{sos}\rangle$ and $\langle$ eos $\rangle$ are special tokens which denote the start and end of a sequence. Also, a special \texttt{[NULL]} token appears when $d_i^k$ is inevitably missing (e.g., font type for a non-text element), or padding variable-length sequences when training a mini-batch.

\begin{equation}
     \mathcal{D}=\left\{\langle\operatorname{sos}\rangle c_1 x_1 y_1 w_1 h_1 t_1 \ldots c_N x_N y_N w_N h_N t_N\langle\operatorname{eos}\rangle\right\}
     \label{eq:1}
\end{equation}

\noindent
\textbf{Representing with Discrete Variables} Following LayoutTransformer~\cite{gupta2021layouttransformer}, we applied an 8-bit uniform quantization on every document element (image region or text) and modelled them using Categorical distribution. We note that while converting coordinates into discrete values leads to some loss of precision, this approach enables the modeling of multiple kinds of distributions, which is crucial for document layouts. Every document object (text or non-text) is projected to the same dimension such that we can concatenate every element $(c_N, x_N, y_N, w_N, h_N, t_N)$ in a single linear sequence of their element values. The overall structure of a page can then be represented by a sequence of $m$ latent vectors where $m$ is decided by the total number of tokens encoded in the input sequence $S$. For conciseness, we use $\boldsymbol{\theta}_j, j \in\{1, \ldots, m\}$ to represent any document element in the above sequence. We model this joint distribution as a product over a series of conditional distributions using the chain rule as shown in Equation~\ref{eq:2}.

\begin{equation}
    p\left(\boldsymbol{\theta}_{1: m}\right)=\prod_{j=1}^{m} p\left(\boldsymbol{\theta}_j \mid \boldsymbol{\theta}_{1: j-1}\right)
    \label{eq:2}
\end{equation}

\subsection{Model Architecture}

\texttt{DocSynthv2} is a document generation transformer pre-trained on document datasets containing multiple elements with a combined set of layout and text attributes. The model learns neural representations of document data, capturing both physical and logical relationships of the document elements with the previously predicted element. Our overall architecture of DocSynthv2 is shown in Figure~\ref{fig:architecture}.

\noindent
\textbf{Training:} Given an initial set of $T$ visible tokens as input containing attributes representing: 1) Layout Category (eg. Table, Table Cell, Paragraph, Title, Caption etc.) 2) Position  3) Font Style 4) Text Content, the model tries to predict the next element with an autoregressive GPT-2 Transformer decoder~\cite{radford2019language}. Each of these GPT blocks consists of a masked multi-head attention (MHA) and a feedforward network (FFN) as shown. The output at the final layer corresponds to the next parameter. 

\noindent
\textbf{Inference:}  During inference, both the position and text tokens are synthesized auto-regressively for the fixed category token (i.e. a reference layout you would like to generate). During both training and inference, the ground-truth sequences have been used to train the model more efficiently as done in ~\cite{gupta2021layouttransformer}.  

\noindent
\textbf{Losses:} Since the model has both continuous and discrete sets of parameters as already discussed, we use a variational loss to minimize KL-Divergence between the softmax predictions for all discrete parameters as in ~\cite{gupta2021layouttransformer}. 





\section{Experiments}\label{sec:experiments}

\begin{figure*}[t]

\centering
\begin{minipage}[c]{0.55\linewidth} 

\includegraphics[width=\linewidth, keepaspectratio]{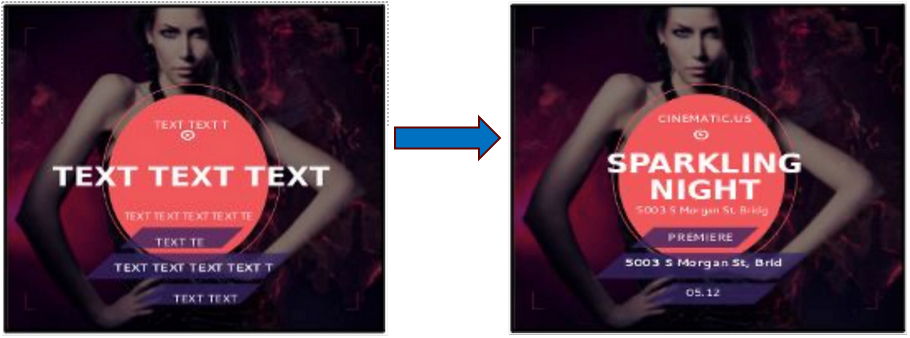} \\
\centering
(a) \\ 
\end{minipage}
\hfill
\begin{minipage}[c]{0.4\linewidth}
\includegraphics[width=\linewidth, keepaspectratio]{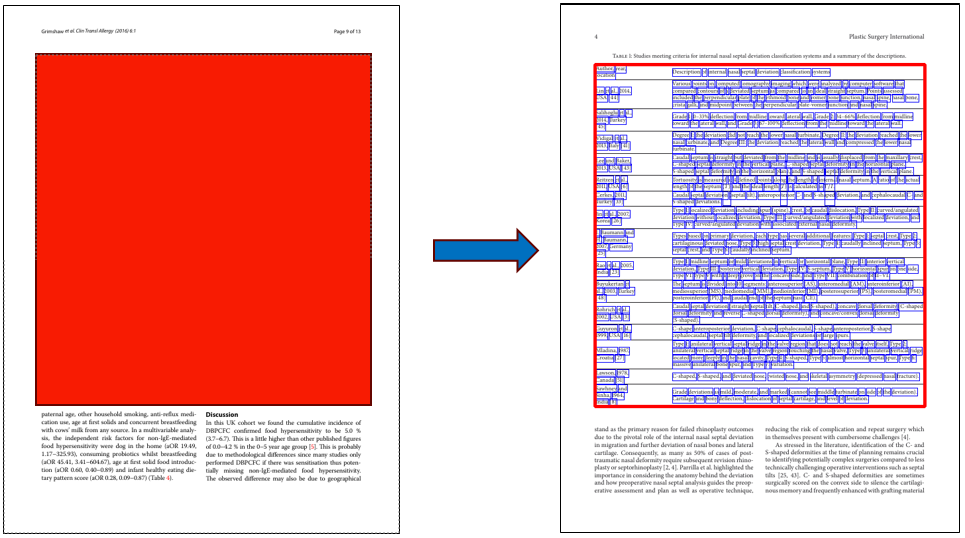} \\
\centering
(b) \\ 
\end{minipage}
\caption{Qualitative results over the two tasks a) \textbf{Text Prediction}: Input (L) and Predicted text content (R) for a Crello sample b) \textbf{Document Completion}: Input (L) with partially filled document and Predicted output (R) for PubGenNet sample}
\label{fig:results}
\end{figure*}

\subsection{Datasets}

Our evaluation of DocSynthv2 primarily utilizes two vector graphic document datasets, Crello~\cite{yamaguchi2021canvasvae} and DocGenNet, our curated version of PubLayNet~\cite{zhong2019publaynet} streamlined for the task of document generation.

\noindent
\textbf{Crello}: Originating from an online design platform, this dataset encompasses a broad range of design templates, including but not limited to social media posts, banner ads, blog headers, and printed materials. We use a similar experimental setting as used in FlexDM~\cite{inoue2023towards}. The released dataset by the authors was partitioned into 18,738 training instances, 2,313 for validation, and 2,271 for testing. Detailed definitions of each attribute can be found in the original paper~\cite{yamaguchi2021canvasvae}. 

\noindent
\textbf{PubGenNet}: For experimental validation, we generated a new benchmark called "PubGenNet," a large-scaled extended dataset curated to advance the field of document generation. This dataset was assembled by extracting a diverse array of samples from the original PubLayNet dataset~\cite{zhong2019publaynet}, which itself is derived from an extensive collection of scientific publications available in PubMed Central. To ensure a comprehensive set of text attributes (eg. font type) along with raw textual content, we utilized a PDF extraction procedure, using the PyMuPDF library enabling us to align this extracted data with the original COCO annotations. In summary, the overall curation process involved extraction and processing of layout and text data from a set of documents represented in the PubLayNet format. After obtaining the document-specific attributes, the processed data was compiled into a structured dataset suitable for training and evaluating document generation models. The resulting training and validation instances are similar to the dataset statistics in PubLayNet with 335,703 document samples for training and 11,245 instances for validation.

\subsection{Tasks}

The primary motivations for our model are to address the key aspects of document design and generation. We have selected the evaluation tasks based on: (1) Creating a new document or completing a partially finished one, focusing on maintaining coherence, appearance, and relevance to the intended content. (2) Test the model's ability in layout design, specifically its understanding of spacing, alignment, and the interplay between text and other elements.\\ 
\noindent
\textbf{Document Completion:} This task requires the model to analyze the current layout elements and content within the document (eg. text, title, tables, figures etc.) and logically predict what elements should follow to maintain the coherence and plausible structure of a document.   

\noindent
\textbf{Single and Multiple Text Box Placement:} This task in terms of next element prediction requires the model to identify optimal locations and sizes for text boxes within a document, based on the existing layout and design principles. It assesses the model's capability to seamlessly incorporate new text elements, ensuring they align with the document's structure and visual appeal.

\subsection{Quantitative Evaluation}

Table~\ref{tab:document} summarizes the performance comparison of DocSynthv2 over the existing SOTA transformer decoder-only models. Our full model (with text attributes) gives us boost in performance over the layout-only model, demonstrating that utilizng the raw text can help guide models for layout generation when avaialble. Although our model is a lightweight decoder-only architecture, it can perform on par with LayoutFormer++~\cite{jiang2023layoutformer++} which is an encoder-decoder-based transformer. Our results with high Alignment and Overlap scores also suggest that \textit{layout generation and completion models gain substantial improvement when trained on sequences integrating textual content}.

In Table~\ref{tab:crello}, we summarize the performance of Single and Multiple Text Box Placement in the Crello dataset. The results show that the model does worse for text placement in the Single Text box condition, likely due to the weaker multimodal features compared to \cite{inoue2023towards}. However, it performs on par for IoU and outperforms for BDE in the Multiple condition, which may be due to the raw text in our model.  

\begin{table}[h]
\begin{tabular}{@{}ccccc@{}}
\toprule
                     & mIoU$\uparrow$           & FID$\downarrow$             & Align$\downarrow$      & Over$\downarrow$         \\ \midrule
LayoutTrans~\cite{gupta2021layouttransformer}    & 0.077          & 14.769          & {\ul 0.019}    & \textbf{0.0013} \\
Layoutformer++~\cite{jiang2023layoutformer++}       & \textbf{0.471} & \textbf{10.251} & 0.020          & 0.0022          \\
Ours (w/o txt) & 0.315          & 12.217          & 0.025          & 0.0019          \\
\textbf{Ours (lay+txt)} & {\ul 0.452}    & {\ul 10.718}    & \textbf{0.015} & \textbf{0.0013} \\
$\Delta$               & \textcolor{red}{-0.019}         & \textcolor{red}{0.467}           & \textcolor{ForestGreen}{-0.004}         & \textcolor{ForestGreen}{0.00}               \\ \bottomrule
\end{tabular}
\caption{Quantitative evaluation for Document Completion. Results style: \textbf{best}, \underline{second best}. $\uparrow$ higher is better and $\downarrow$ lower is better}
\label{tab:document}
\end{table}

\begin{table}[h]
\begin{tabular}{@{}lcccc@{}}
\toprule
                 & \multicolumn{2}{c}{Single}      & \multicolumn{2}{c}{Multiple}    \\ \midrule
                 & IoU$\uparrow$             & BDE$\downarrow$            & IoU$\uparrow$             & BDE$\downarrow$            \\
SmartText~\cite{li2021harmonious}        & 0.047          & 0.262          & 0.023          & 0.300          \\
FlexDM (MM)~\cite{inoue2023towards}      & \textbf{0.357} & \textbf{0.098} & \textbf{0.110} & 0.141          \\
FlexDM (w/o img)~\cite{inoue2023towards} & {\ul 0.355}    & {\ul 0.100}    & {\ul 0.103}    & 0.156          \\
FlexDM (w/o txt)~\cite{inoue2023towards} & 0.350          & 0.106          & 0.086          & 0.178          \\
\textbf{Ours}             & 0.315          & 0.104          & 0.105          & \textbf{0.131} \\ \bottomrule
\end{tabular}
\caption{Quantitaive evaluation for Single and Multiple Box Placement in Crello. Results style: \textbf{best}, \underline{second best}. $\uparrow$ higher is better and $\downarrow$ lower is better}
\label{tab:crello}
\end{table}

\subsection{Qualitative Evaluation}

Figure \ref{fig:results} shows example of our  applied for text synthesis and document completion on the Crello and PubGenNet datasets. In the Crello Text prediction example, it can be seen that the text is aligned with the layout showing a plausible flyer title  for the heading section followed by an address and date in the sub text fields. For the Document Completion Task, we have the model generate the text within in an existing Table structure. The filled text maintains coherence across the two table columns, filling it with Authors names and reference information on the left and text of the right. In this example the text coherence could likely be improved by LLMs.

\section{Future Scope and Challenges}

In conclusion, DocSynthv2 demonstrates that integrating text with layout sequences into an autoregressive framework enriches the data representation and provides additional context, leading to improved stability and performance in generating coherent and contextually appropriate document content and motivates future work.
 First, the integration of layout and text needs to advance beyond current capabilities to \textit{address the diversity of document styles and industry-specific standards}. We believe future work may benefit from visual-language models~\cite{tang2023unifying} that can understand multimodal content or code generation models~\cite{tang2023layoutnuwa} that can learn  complex structure from a wide array of document formats and content types. We also believe, the evaluation of document generation systems remains a critical challenge. There is a pressing need for \textit{evaluation frameworks which can effectively measure the usefulness of generated documents} in terms of both their visual layout and textual content. These frameworks must encompass metrics that evaluate coherence, relevance, readability, and visual appeal, reflecting the multi-functional nature of documents. 




\section*{Acknowledgement}
The resources and support from the Adobe Document Intelligence Lab (DIL) team were instrumental in the successful completion of this project. Special thanks to Ani Nenkova, Joe Barrow, Varun Manjunatha and Chris Tensmeyer, whose guidance and expertise were invaluable throughout the internship. Additionally, Sanket Biswas expresses his gratitude to Nora Graichen for her constant assistance and perceptive criticism, particularly during the last phases of submission.

{
    \small
    \bibliographystyle{ieeenat_fullname}
    \bibliography{main}

\begin{thebibliography}{35}
\providecommand{\natexlab}[1]{#1}
\providecommand{\url}[1]{\texttt{#1}}
\expandafter\ifx\csname urlstyle\endcsname\relax
  \providecommand{\doi}[1]{doi: #1}\else
  \providecommand{\doi}{doi: \begingroup \urlstyle{rm}\Url}\fi

\bibitem[Arroyo et~al.(2021)Arroyo, Postels, and Tombari]{arroyo2021variational}
Diego~Martin Arroyo, Janis Postels, and Federico Tombari.
\newblock Variational transformer networks for layout generation.
\newblock In \emph{Proceedings of the IEEE/CVF Conference on Computer Vision and Pattern Recognition}, pages 13642--13652, 2021.

\bibitem[Biswas et~al.(2021)Biswas, Riba, Llad{\'o}s, and Pal]{biswas2021docsynth}
Sanket Biswas, Pau Riba, Josep Llad{\'o}s, and Umapada Pal.
\newblock Docsynth: a layout guided approach for controllable document image synthesis.
\newblock In \emph{Document Analysis and Recognition--ICDAR 2021: 16th International Conference, Lausanne, Switzerland, September 5--10, 2021, Proceedings, Part III}, pages 555--568. Springer, 2021.

\bibitem[Brown et~al.(2020)Brown, Mann, Ryder, Subbiah, Kaplan, Dhariwal, Neelakantan, Shyam, Sastry, Askell, et~al.]{brown2020language}
Tom Brown, Benjamin Mann, Nick Ryder, Melanie Subbiah, Jared~D Kaplan, Prafulla Dhariwal, Arvind Neelakantan, Pranav Shyam, Girish Sastry, Amanda Askell, et~al.
\newblock Language models are few-shot learners.
\newblock \emph{Advances in neural information processing systems}, 33:\penalty0 1877--1901, 2020.

\bibitem[Deka et~al.(2017)Deka, Huang, Franzen, Hibschman, Afergan, Li, Nichols, and Kumar]{deka2017rico}
Biplab Deka, Zifeng Huang, Chad Franzen, Joshua Hibschman, Daniel Afergan, Yang Li, Jeffrey Nichols, and Ranjitha Kumar.
\newblock Rico: A mobile app dataset for building data-driven design applications.
\newblock In \emph{Proceedings of the 30th annual ACM symposium on user interface software and technology}, pages 845--854, 2017.

\bibitem[Esser et~al.(2021)Esser, Rombach, and Ommer]{esser2021taming}
Patrick Esser, Robin Rombach, and Bjorn Ommer.
\newblock Taming transformers for high-resolution image synthesis.
\newblock In \emph{Proceedings of the IEEE/CVF conference on computer vision and pattern recognition}, pages 12873--12883, 2021.

\bibitem[Feng et~al.(2024)Feng, Zhu, Fu, Jampani, Akula, He, Basu, Wang, and Wang]{feng2024layoutgpt}
Weixi Feng, Wanrong Zhu, Tsu-jui Fu, Varun Jampani, Arjun Akula, Xuehai He, Sugato Basu, Xin~Eric Wang, and William~Yang Wang.
\newblock Layoutgpt: Compositional visual planning and generation with large language models.
\newblock \emph{Advances in Neural Information Processing Systems}, 36, 2024.

\bibitem[Gupta et~al.(2021)Gupta, Lazarow, Achille, Davis, Mahadevan, and Shrivastava]{gupta2021layouttransformer}
Kamal Gupta, Justin Lazarow, Alessandro Achille, Larry~S Davis, Vijay Mahadevan, and Abhinav Shrivastava.
\newblock Layouttransformer: Layout generation and completion with self-attention.
\newblock In \emph{Proceedings of the IEEE/CVF International Conference on Computer Vision}, pages 1004--1014, 2021.

\bibitem[He et~al.(2021)He, Liao, Yang, Yang, Song, Rosenhahn, and Xiang]{he2021context}
Sen He, Wentong Liao, Michael~Ying Yang, Yongxin Yang, Yi-Zhe Song, Bodo Rosenhahn, and Tao Xiang.
\newblock Context-aware layout to image generation with enhanced object appearance.
\newblock In \emph{Proceedings of the IEEE/CVF conference on computer vision and pattern recognition}, pages 15049--15058, 2021.

\bibitem[Hui et~al.(2023)Hui, Zhang, Zhang, Xie, Wang, and Lu]{hui2023unifying}
Mude Hui, Zhizheng Zhang, Xiaoyi Zhang, Wenxuan Xie, Yuwang Wang, and Yan Lu.
\newblock Unifying layout generation with a decoupled diffusion model.
\newblock In \emph{Proceedings of the IEEE/CVF Conference on Computer Vision and Pattern Recognition}, pages 1942--1951, 2023.

\bibitem[Inoue et~al.(2023{\natexlab{a}})Inoue, Kikuchi, Simo-Serra, Otani, and Yamaguchi]{inoue2023layoutdm}
Naoto Inoue, Kotaro Kikuchi, Edgar Simo-Serra, Mayu Otani, and Kota Yamaguchi.
\newblock Layoutdm: Discrete diffusion model for controllable layout generation.
\newblock In \emph{Proceedings of the IEEE/CVF Conference on Computer Vision and Pattern Recognition}, pages 10167--10176, 2023{\natexlab{a}}.

\bibitem[Inoue et~al.(2023{\natexlab{b}})Inoue, Kikuchi, Simo-Serra, Otani, and Yamaguchi]{inoue2023towards}
Naoto Inoue, Kotaro Kikuchi, Edgar Simo-Serra, Mayu Otani, and Kota Yamaguchi.
\newblock Towards flexible multi-modal document models.
\newblock In \emph{Proceedings of the IEEE/CVF Conference on Computer Vision and Pattern Recognition}, pages 14287--14296, 2023{\natexlab{b}}.

\bibitem[Jiang et~al.(2023)Jiang, Guo, Sun, Deng, Wu, Mijovic, Yang, Lou, and Zhang]{jiang2023layoutformer++}
Zhaoyun Jiang, Jiaqi Guo, Shizhao Sun, Huayu Deng, Zhongkai Wu, Vuksan Mijovic, Zijiang~James Yang, Jian-Guang Lou, and Dongmei Zhang.
\newblock Layoutformer++: Conditional graphic layout generation via constraint serialization and decoding space restriction.
\newblock In \emph{Proceedings of the IEEE/CVF Conference on Computer Vision and Pattern Recognition}, pages 18403--18412, 2023.

\bibitem[Johnson et~al.(2018)Johnson, Gupta, and Fei-Fei]{johnson2018image}
Justin Johnson, Agrim Gupta, and Li Fei-Fei.
\newblock Image generation from scene graphs.
\newblock In \emph{Proceedings of the IEEE conference on computer vision and pattern recognition}, pages 1219--1228, 2018.

\bibitem[Jyothi et~al.(2019)Jyothi, Durand, He, Sigal, and Mori]{jyothi2019layoutvae}
Akash~Abdu Jyothi, Thibaut Durand, Jiawei He, Leonid Sigal, and Greg Mori.
\newblock Layoutvae: Stochastic scene layout generation from a label set.
\newblock In \emph{Proceedings of the IEEE/CVF International Conference on Computer Vision}, pages 9895--9904, 2019.

\bibitem[Kim et~al.(2022)Kim, Hong, Yim, Nam, Park, Yim, Hwang, Yun, Han, and Park]{kim2022ocr}
Geewook Kim, Teakgyu Hong, Moonbin Yim, JeongYeon Nam, Jinyoung Park, Jinyeong Yim, Wonseok Hwang, Sangdoo Yun, Dongyoon Han, and Seunghyun Park.
\newblock Ocr-free document understanding transformer.
\newblock In \emph{European Conference on Computer Vision}, pages 498--517. Springer, 2022.

\bibitem[Kong et~al.(2022)Kong, Jiang, Chang, Zhang, Hao, Gong, and Essa]{kong2022blt}
Xiang Kong, Lu Jiang, Huiwen Chang, Han Zhang, Yuan Hao, Haifeng Gong, and Irfan Essa.
\newblock Blt: Bidirectional layout transformer for controllable layout generation.
\newblock In \emph{European Conference on Computer Vision}, pages 474--490. Springer, 2022.

\bibitem[Li et~al.(2021)Li, Zhang, and Wang]{li2021harmonious}
Chenhui Li, Peiying Zhang, and Changbo Wang.
\newblock Harmonious textual layout generation over natural images via deep aesthetics learning.
\newblock \emph{IEEE Transactions on Multimedia}, 24:\penalty0 3416--3428, 2021.

\bibitem[Li et~al.(2019)Li, Yang, Hertzmann, Zhang, and Xu]{li2019layoutgan}
Jianan Li, Jimei Yang, Aaron Hertzmann, Jianming Zhang, and Tingfa Xu.
\newblock Layoutgan: Generating graphic layouts with wireframe discriminators.
\newblock \emph{arXiv preprint arXiv:1901.06767}, 2019.

\bibitem[Patil et~al.(2020)Patil, Ben-Eliezer, Perel, and Averbuch-Elor]{patil2020read}
Akshay~Gadi Patil, Omri Ben-Eliezer, Or Perel, and Hadar Averbuch-Elor.
\newblock Read: Recursive autoencoders for document layout generation.
\newblock In \emph{Proceedings of the IEEE/CVF Conference on Computer Vision and Pattern Recognition Workshops}, pages 544--545, 2020.

\bibitem[Patil et~al.(2021)Patil, Li, Fisher, Savva, and Zhang]{patil2021layoutgmn}
Akshay~Gadi Patil, Manyi Li, Matthew Fisher, Manolis Savva, and Hao Zhang.
\newblock Layoutgmn: Neural graph matching for structural layout similarity.
\newblock In \emph{Proceedings of the IEEE/CVF Conference on Computer Vision and Pattern Recognition}, pages 11048--11057, 2021.

\bibitem[Pfitzmann et~al.(2022)Pfitzmann, Auer, Dolfi, Nassar, and Staar]{pfitzmann2022doclaynet}
Birgit Pfitzmann, Christoph Auer, Michele Dolfi, Ahmed~S Nassar, and Peter Staar.
\newblock Doclaynet: a large human-annotated dataset for document-layout segmentation.
\newblock In \emph{Proceedings of the 28th ACM SIGKDD Conference on Knowledge Discovery and Data Mining}, pages 3743--3751, 2022.

\bibitem[Pisaneschi et~al.(2023)Pisaneschi, Gemelli, and Marinai]{pisaneschi2023automatic}
Lorenzo Pisaneschi, Andrea Gemelli, and Simone Marinai.
\newblock Automatic generation of scientific papers for data augmentation in document layout analysis.
\newblock \emph{Pattern Recognition Letters}, 167:\penalty0 38--44, 2023.

\bibitem[Radford et~al.(2019)Radford, Wu, Child, Luan, Amodei, Sutskever, et~al.]{radford2019language}
Alec Radford, Jeffrey Wu, Rewon Child, David Luan, Dario Amodei, Ilya Sutskever, et~al.
\newblock Language models are unsupervised multitask learners.
\newblock \emph{OpenAI blog}, 1\penalty0 (8):\penalty0 9, 2019.

\bibitem[Shimoda et~al.(2024)Shimoda, Haraguchi, Uchida, and Yamaguchi]{shimoda2024towards}
Wataru Shimoda, Daichi Haraguchi, Seiichi Uchida, and Kota Yamaguchi.
\newblock Towards diverse and consistent typography generation.
\newblock In \emph{Proceedings of the IEEE/CVF Winter Conference on Applications of Computer Vision}, pages 7296--7305, 2024.

\bibitem[Tang et~al.(2023{\natexlab{a}})Tang, Wu, Li, and Duan]{tang2023layoutnuwa}
Zecheng Tang, Chenfei Wu, Juntao Li, and Nan Duan.
\newblock Layoutnuwa: Revealing the hidden layout expertise of large language models.
\newblock \emph{arXiv preprint arXiv:2309.09506}, 2023{\natexlab{a}}.

\bibitem[Tang et~al.(2023{\natexlab{b}})Tang, Yang, Wang, Fang, Liu, Zhu, Zeng, Zhang, and Bansal]{tang2023unifying}
Zineng Tang, Ziyi Yang, Guoxin Wang, Yuwei Fang, Yang Liu, Chenguang Zhu, Michael Zeng, Cha Zhang, and Mohit Bansal.
\newblock Unifying vision, text, and layout for universal document processing.
\newblock In \emph{Proceedings of the IEEE/CVF Conference on Computer Vision and Pattern Recognition}, pages 19254--19264, 2023{\natexlab{b}}.

\bibitem[Touvron et~al.(2023)Touvron, Lavril, Izacard, Martinet, Lachaux, Lacroix, Rozi{\`e}re, Goyal, Hambro, Azhar, et~al.]{touvron2023llama}
Hugo Touvron, Thibaut Lavril, Gautier Izacard, Xavier Martinet, Marie-Anne Lachaux, Timoth{\'e}e Lacroix, Baptiste Rozi{\`e}re, Naman Goyal, Eric Hambro, Faisal Azhar, et~al.
\newblock Llama: Open and efficient foundation language models.
\newblock \emph{arXiv preprint arXiv:2302.13971}, 2023.

\bibitem[Vaswani et~al.(2017)Vaswani, Shazeer, Parmar, Uszkoreit, Jones, Gomez, Kaiser, and Polosukhin]{vaswani2017attention}
Ashish Vaswani, Noam Shazeer, Niki Parmar, Jakob Uszkoreit, Llion Jones, Aidan~N Gomez, {\L}ukasz Kaiser, and Illia Polosukhin.
\newblock Attention is all you need.
\newblock \emph{Advances in neural information processing systems}, 30, 2017.

\bibitem[Yamaguchi(2021)]{yamaguchi2021canvasvae}
Kota Yamaguchi.
\newblock Canvasvae: Learning to generate vector graphic documents.
\newblock In \emph{Proceedings of the IEEE/CVF International Conference on Computer Vision}, pages 5481--5489, 2021.

\bibitem[Yim et~al.(2021)Yim, Kim, Cho, and Park]{yim2021synthtiger}
Moonbin Yim, Yoonsik Kim, Han-Cheol Cho, and Sungrae Park.
\newblock Synthtiger: Synthetic text image generator towards better text recognition models.
\newblock In \emph{International conference on document analysis and recognition}, pages 109--124. Springer, 2021.

\bibitem[Zhang et~al.(2023)Zhang, Guo, Sun, Lou, and Zhang]{zhang2023layoutdiffusion}
Junyi Zhang, Jiaqi Guo, Shizhao Sun, Jian-Guang Lou, and Dongmei Zhang.
\newblock Layoutdiffusion: Improving graphic layout generation by discrete diffusion probabilistic models.
\newblock In \emph{Proceedings of the IEEE/CVF International Conference on Computer Vision}, pages 7226--7236, 2023.

\bibitem[Zhao et~al.(2020)Zhao, Yin, Meng, and Sigal]{zhao2020layout2image}
Bo Zhao, Weidong Yin, Lili Meng, and Leonid Sigal.
\newblock Layout2image: Image generation from layout.
\newblock \emph{International Journal of Computer Vision}, 128\penalty0 (10):\penalty0 2418--2435, 2020.

\bibitem[Zheng et~al.(2023)Zheng, Zhou, Li, Qi, Shan, and Li]{zheng2023layoutdiffusion}
Guangcong Zheng, Xianpan Zhou, Xuewei Li, Zhongang Qi, Ying Shan, and Xi Li.
\newblock Layoutdiffusion: Controllable diffusion model for layout-to-image generation.
\newblock In \emph{Proceedings of the IEEE/CVF Conference on Computer Vision and Pattern Recognition}, pages 22490--22499, 2023.

\bibitem[Zheng et~al.(2019)Zheng, Qiao, Cao, and Lau]{zheng2019content}
Xinru Zheng, Xiaotian Qiao, Ying Cao, and Rynson~WH Lau.
\newblock Content-aware generative modeling of graphic design layouts.
\newblock \emph{ACM Transactions on Graphics (TOG)}, 38\penalty0 (4):\penalty0 1--15, 2019.

\bibitem[Zhong et~al.(2019)Zhong, Tang, and Yepes]{zhong2019publaynet}
Xu Zhong, Jianbin Tang, and Antonio~Jimeno Yepes.
\newblock Publaynet: largest dataset ever for document layout analysis.
\newblock In \emph{2019 International conference on document analysis and recognition (ICDAR)}, pages 1015--1022. IEEE, 2019.

\end{thebibliography}
}


\end{document}